# From Ad Hoc Pilots to Repeatable Patterns: Structuring Drone Collaboration in Emergency Services with *DroneLets*

*Completed Research Paper*

**Dzmitry Katsiuba**
University of Zurich
Zurich, Switzerland
katsiuba@ifi.uzh.ch

**Samuel Brander**
University of Zurich
Zurich, Switzerland
samuel.brander@uzh.ch

**Mateusz Dolata**
Zeppelin University
Friedrichshafen, Germany
mateusz.dolata@zu.de

**Gerhard Schwabe**
University of Zurich
Zurich, Switzerland
schwabe@ifi.uzh.ch

## Abstract

*Drones hold promise for supporting emergency services, but their integration into workflows remains ad hoc and coordination-intensive. This paper addresses two research questions: how emergency teams want to collaborate with drones, and how to formalize these collaborations into repeatable processes. Based on four field trials and 95 interviews, we derive 44 interaction patterns grouped into 10 meta-patterns reflecting operational needs such as reconnaissance, communication, and logistical support. To structure these practices, we introduce DroneLets – a new class of design artifacts that extend Collaboration Engineering to embodied agents. DroneLets capture setup requirements, drone capabilities, environmental constraints, and coordinated actions across human and drone actors. They offer a modular framework for designing repeatable, scalable collaboration processes in emergency services, illustrated through patterns such as broadcasting to bystanders and post-fire monitoring. This work expands the scope of CE and provides a structured foundation for integrating autonomous drones into high-stakes field operations.*

## Introduction

The integration of drones into emergency services (ES) holds great potential for improving situational awareness, speeding up response times, and increasing the safety of human responders (Wankmüller et al., 2021). However, embedding drones into real-life emergency workflows remains a significant organizational and technical challenge. Today, the deployment of drones typically depends on the presence of a dedicated drone pilot. This pilot assumes a role analogous to that of a facilitator in collaborative work – controlling the drone, interpreting information, and ensuring that the drone's contributions align with the team's goals and procedures. The industry, however, is progressing along a clear spectrum of autonomy. This ranges from semi-autonomous systems, which execute pre-programmed missions under supervision, to highly or fully autonomous drones capable of independent navigation and decision-making in complex and even GPS-denied environments. As drone technology becomes increasingly autonomous, the question arises: how can we go beyond manual control and systematically integrate autonomous drones into ES workflows?

The successful integration of autonomous drones into ES hinges on both technological advancements and the development of suitable organizational structures and collaborative processes. While technological autonomy – the ability of drones to operate independently – is limited by the maturity of sensor technologies, navigation systems, and intelligent decision-making, organizational autonomy – the ability to utilize drones effectively within human workflows – requires consideration of specific operational scenarios and the unique, domain-specific needs of emergency service teams.

To address this challenge, we draw on the theory and practices of Collaboration Engineering (CE) – a discipline that provides structured methods for designing repeatable and effective collaboration processes (Briggs et al., 2003). CE has traditionally focused on human-human collaboration, offering constructs such as ThinkLets to capture and transfer facilitation knowledge. Over time, these traditional collaboration settings were augmented by digital tools and platforms, which enabled the structured execution of ThinkLets in distributed, technologically supported environments. More recently, thanks to advancements in artificial intelligence (AI), particularly the emergence of Large Language Models (LLMs), the scope of CE has begun to shift toward hybrid collaboration, where AI-based agents are integrated as participants in collaborative processes. This evolution has led to the development of AI-ThinkLets (Schwabe et al., 2025), which adapt CE's principles to include digital actors. At the same time, rapid advancements in robotics and embodied AI are setting the stage for the next challenge in designing collaborative processes.

However, the transfer from digital to embodied intelligent systems, such as robots and autonomous drones, represents qualitatively different challenges in ES. Unlike LLMs or digital brainstorming agents, drones operate in physical space, interact with dynamic and potentially hazardous environments, and require spatial-temporal coordination with human actors. While prior work has shown that CE can accommodate digital augmentation in cognitive collaboration (e.g., ideation), the collaboration between human actors and embodied, agentic systems remains underexplored in the CE literature. Existing CE concepts are hard to apply in context of embodied agents like drones due to the typical spatial and safety challenges, which this study addresses by proposing a new class of design constructs that account for physical coordination, environmental awareness, and real-time decision-making in high-stakes settings.

For drones to be effectively integrated into ES operations, both technological advancements and organizational changes must occur simultaneously. Technological progress should be accompanied by systematic approaches for designing and managing collaboration between humans and drones. To address this challenge, this study proceeds in two steps. We first investigate *how emergency teams envision effective collaboration with drones* (RQ1). Building on these insights, we look at *how we can provide human drone pilots with repeatable collaboration processes for emergency services* (RQ2).

To address RQ1, drawing on field trial data and practitioner insights, we derive a Drone Collaboration Pattern Catalogue, identifying recurring scenarios and functions that autonomous drones can fulfill within ES operations. To address RQ2, we propose a novel construct – DroneLets. We define a *DroneLet* as a named and packaged facilitation recipe that enables a reliable, reproducible pattern of collaboration between human responders and autonomous drones in the physical world. Built on the ThinkLet logic, each DroneLet specifies when the pattern should or should not be applied, the required physical and digital environment, the drone's capabilities and appearance, the setup steps for both the setting and the aircraft, and the scripted actions for human actors and the drone, including error-handling rules, so that emergency-service teams can instate the same structured human-drone workflow, with consistent outcomes, across incidents. that captures repeatable patterns of human-drone collaboration in emergency contexts and provides a design structure for embedding autonomous drones into coordinated human-drone workflows. Together, these contributions extend the scope of CE to accommodate embodied, agentic systems, opening new opportunities for research and practice in collaborative work design.

DroneLets emerged in a staged process with two intermediate conceptualization steps. We first derived *interaction patterns*, which were then grouped these into 10 higher-level *meta-patterns* based on their shared collaborative purpose. In the context of this article, an *interaction pattern* is a concise, field-derived description of a recurring collaborative action sequence in which humans and drones coordinate, communicate, and cooperate to accomplish a specific task during emergency operations. A meta-pattern is an overarching category that clusters multiple interaction patterns based on their shared collaborative purpose, summarizing how recurring human-drone coordination, communication, and cooperation strategies address a common operational need in emergency services. Whereas meta-patterns identify and organize the common goals underlying multiple interaction patterns, DroneLets transform a chosen meta-pattern into a fully scripted, executable collaborative procedure for human-drone teams that delivers predictable outcomes when applied across contexts (e.g., in search and rescue, during firefighting, and after a shooting).

While the collaborative patterns we propose can enhance coordination in various scenarios involving human pilots or drones acting fully independently, our primary focus is on structuring collaboration for systems that can act autonomously based on commands from a drone officer who also acts as the next escalation level to handle exceptions. The goal is to provide a formal methodology that enables emergency services to move beyond the dependency on a dedicated, hands-on pilot, and instead transition towards a model where a single operator can supervise the missions of multiple autonomous assets.

## Related Work

### *Problem: Drone Use in Emergency Services*

Drones have gained increasing attention in recent years for their potential to enhance the effectiveness and responsiveness of emergency services, particularly in policing and firefighting. Their speed, agility, and ability to access hard-to-reach areas offer distinct advantages over traditional ground or aerial vehicles (Boonyard et al., 2025). In policing, drones are used for traffic surveillance (J. Lee et al., 2015), search-and-rescue missions (García-Hernández et al., 2019) and documenting crime scenes from the air (Urbanová et al., 2017). In firefighting, drones can provide aerial perspectives of large fires (Wankmüller et al., 2021), access high-rise buildings or remote forest areas, and assist in identifying hotspots using thermal imaging and even three-dimensional modeling and damage assessment (Zweglinski, 2020).

Numerous studies have highlighted the operational benefits of drone deployments in first responder contexts. Researchers emphasize that drones can arrive on-site faster than traditional emergency vehicles, thereby reducing response times (Desjardins et al., 2014; Mayer et al., 2019), reduce operational risks in rescue missions, thereby increasing the safety and performance of response teams (Wankmüller et al., 2021). Heen et al. (2018) point out that drones offer more maneuverability, while also being more cost-effective than helicopters. In firefighting scenarios, drones have proven valuable for accessing hazardous or inaccessible environments, such as upper floors of skyscrapers or dense wildfire areas (Wankmüller et al., 2021). Thermal cameras attached to drones can detect fire hotspots, and future applications may even include automated extinguishing capabilities, such as drones dropping fire suppressant balls from the air (Aydin et al., 2019; Soliman et al., 2019).

Despite these benefits, current drone deployments in ES heavily rely on human drone pilots, who must manually operate the device, interpret the video feeds, and relay relevant information to the rest of the team. This model burdens the pilot, who often functions as a technical operator and informal facilitator, mediating between the drone and other team members. Early studies indicate that integrating autonomous drones into emergency services may pose challenges beyond information overload. These challenges could potentially lead to changes in work arrangements (Li et al., 2024). Thus, effective deployment of drones in complex environments requires tight coordination, situational awareness, and communication, often under high time pressure.

The emergence of autonomous drones promises to reduce the cognitive and operational load on human pilots, enabling drones to navigate, observe, and report independently. However, this increased autonomy introduces new challenges related to trust, privacy (Dolata & Schwabe, 2023), safety and feeling of safety (Wu et al., 2025), control delegation, accountability, and synchronization with human teams (Li et al., 2024). Unlike remotely piloted systems, autonomous drones make real-time decisions based on environmental inputs, raising critical questions: How do team members know what the drone is doing? How should they interact with it? This paper addresses this gap by exploring how structured collaboration approaches can be adapted to support the integration of autonomous drones in high-stakes environments.

Traditionally, collaboration and interaction between various entities during an emergency have relied on scripts known as standard operating procedures (SOPs). These established protocols are designed to ensure that personnel can efficiently execute tasks and make informed decisions, even in high-pressure environments, while acting safely and predictably (Butler et al., 2023; Duncan et al., 2014). Prior research indicates that SOPs effectively support firefighters and police in most situations they encounter (Weinschenk et al., 2008) and provide a stable basis for reference, even in non-standard situations (Carvalho et al., 2018; Jahn, 2016). Accordingly, training often focuses on internalizing these procedures, enabling responders to react instinctively and correctly during emergencies (Toups Dugas & Kerne, 2007; Weinschenk et al., 2008).

SOPs are frequently documented at the level of specific departments or regions. Their form and content might differ, but they typically describe the scope of a procedure, basic concepts and roles along with their duties during an operation, important considerations about the context (e.g., presence of an elevator), communication and interaction protocols, and actions to be taken. Some are highly specific including exemplary commands while others remain more generic (e.g., (*Fire Department and EMS Standard Operating Procedures*, 2025; *Standard Operating Procedures (SOP)*, 2025)). However, despite the increasing role and great potential of autonomous drones in firefighting and policing, there is a lack of scripts describing how to effectively incorporate such autonomous technology into emergency situations. While some departments attempt to address this gap, the resulting SOPs remain highly abstract and focus on piloted drones. They tend to describe general rules and limitations concerning the use of drones instead of focusing on how collaboration between human and non-human agents in a specific emergency scenario can be established. Furthermore, instead of integrating drone use into existing procedures, such as those for wildlife fires or high-building fires, they often create separate SOPs specifically for drone use (e.g., Unmanned Aircraft Systems (UAS) Standard Operating Procedures (Compliance 1, 2024), Unmanned Aircraft Systems Operations (Russellville Fire Department Policy Manual, 2021)).

### *Solution approach: Collaboration Engineering*

Collaboration Engineering provides a promising lens for structuring collaboration between humans and intelligent systems in complex, high-stakes environments. CE aims to design repeatable and transferable collaboration processes that practitioners can apply without the need for a trained facilitator (Briggs et al., 2003). Collaborative technologies, however, can only be fully realized when integrated into a larger collaboration system. As Briggs (2009) defined it, a collaboration system comprises a diverse group of actors, such as team members and facilitators, along with various tools, including hardware and software, and procedural knowledge that facilitates groups in achieving their objectives effectively and efficiently. Within this framework, the procedural knowledge required for specific collaboration tasks is captured in ThinkLets.

ThinkLet is a modular facilitation technique that guides groups through common collaboration tasks (Briggs & De Vreede, 2009). In other words, it is a named and packaged facilitation intervention or recipe that helps instantiate a known, reliable, and replicable pattern of collaboration among people who work together toward a shared goal (Kolfschoten et al., 2006). Each ThinkLet consists of three main components: the *tools* used (e.g., paper, software, workspaces), the *setup* of the environment and roles, and the *script* that guides participants through the collaboration process. This structured decomposition ensures that collaboration techniques can be repeatedly applied across contexts with predictable outcomes (Briggs & De Vreede, 2009).

Although ThinkLets were originally developed for cognitive group tasks such as idea generation, prioritization, and consensus-building, subsequent research has shown their adaptability to broader collaborative contexts. For example, ThinkLets have been successfully reused and adapted in advisory services (Schwabe et al., 2016), where they supported structured dialogues between advisors and clients. This demonstrated that ThinkLets could support not only cognitive collaboration within groups but also structured, interaction-driven workflows in professional service environments. With the rise of intelligent systems, CE has started to evolve (Memmert & Bittner, 2024; Schwabe et al., 2025). Schwabe et al. (2025) extend the CE framework by introducing AI-ThinkLets, adapting ThinkLet-based collaboration patterns to include digital agents. These AI-ThinkLets were instantiated using GPT-4, a LLM acting as a text-based participant in a brainstorming environment. This marked a significant step in rethinking CE to include artificial actors with generative capabilities. AI-ThinkLets build on the same logic as traditional ThinkLets but provide additional configurations and scripts tailored to the capabilities and limitations of AI agents.

Yet, digital agents like LLM-based chats remain purely virtual – they do not interact with the physical world, nor do they operate in embodied, real-time, and safety-critical environments. Integrating such agents, like autonomous drones, into ES workflows introduces new challenges that exceed the scope of current CE practices. Drones operate in physical space, require coordination with human actors across time and distance, and contribute to mission-critical decision-making under uncertainty. These characteristics raise important questions regarding autonomy levels, interaction timing, spatial awareness, and shared control, none of which are addressed by existing CE constructs.

While AI-ThinkLets demonstrate that collaboration processes can be structured to include virtual agents, a further conceptual expansion is necessary to accommodate embodied intelligent agents operating in real-world contexts. Structuring collaboration with such agents requires new types of setup configurations, interaction scripts that can address not only communication flows but also interaction, behavior of actors involved. This represents a clear theoretical and practical gap – one that calls for an extension of CE principles to account for physical coordination, real-time environmental awareness, and hybrid role delegation between humans and autonomous systems. We argue that doing so requires a new class of collaboration design constructs – an issue we address in the discussion section of this paper.

# Methodology

## *Deriving Interaction Patterns from Empirical Data*

This study draws on empirical insights from four field trials and provides an overview of how autonomous drones can support emergency services. These trials were conducted in the period of 2020-2023 in collaboration with Swiss emergency organizations and involved autonomous drones in realistic emergency service scenarios, including two field trials from each firefighting and policing context. During the field trials, we simulated critical situations such as wildfire and building fire response in collaboration with the fire departments of Silenen and Altdorf, as well as tactical and search-and-rescue police operations conducted in cooperation with the Zurich City Police. In these trials, we explored various modes of drone integration, ranging from operation via a dedicated drone pilot to scenarios involving autonomous drones supported by a drone officer on site. Each trial involved active participation by human teams and autonomous drones and was designed to investigate coordination and operational dynamics, as well as collaboration with bystanders and, depending on the scenario, victims or perpetrators.

Before and after the field trials, we conducted in total of 95 interviews with 100 participants – including 28 firefighters (ranging from frontline to incident commanders), 19 police officers (tactical and patrol), and 53 bystanders who were acquired through broad yet unsystematic convenience sampling to ensure a diverse range of perspectives. The interviews addressed topics such as expectations, situational roles, decision-making processes, and experiences of human-drone interaction. Specifically, we asked participants about practical aspects such as when they would or would not use a drone, in which way it shall interact with various stakeholders and how this interaction could be structured. For the present study, we systematically analyzed this dataset to gain a deeper understanding of how autonomous drones influence collaboration on the ground. For this, we followed a qualitative coding approach based on the methodology by Saldaña (2021). All interview transcripts underwent an iterative coding process, employing both deductive and inductive methodologies. During the coding phase, we systematically evaluated the advantages and disadvantages of drones, analyzed prevailing public attitudes towards drones, identified the design specifications for drone applications, and explored the intricacies of human-drone collaboration. However, our primary focus was on identifying distinct real-life or potential collaboration scenarios.

The identified collaboration scenarios (n=56) formed the empirical foundation of this study. These scenarios describe observed or imagined situations in which humans and drones interact during emergency response. In the next analytical step, we grouped similar scenarios into 44 distinct Interaction Patterns, consolidating cases that described essentially the same collaborative logic, but applied across different operational situations. These patterns capture real-world roles (e.g., drone as scout, drone as messenger, drone as delivery) and coordination flows.

To further structure our findings, we organized a sense-making workshop to analyze the core activities associated with each pattern and to group them. This workshop was conducted with 4 researchers with combined expertise in CE, HCI, and information systems to ensure the theoretical robustness of the resulting meta-patterns. During the workshop, each of the 44 Interaction Patterns was written on a card. The team then engaged in an iterative, open card-sorting exercise, grouping patterns based on their underlying collaborative function and purpose (e.g., "gathering information" vs. "providing support"). This process, which involved multiple rounds of grouping, debating, and refining categories, allowed us to consolidate the patterns into ten overarching Meta-Patterns, reflecting recurring collaboration strategies that characterize human-drone or multi-drone interactions across emergency services. The selection of researchers with deep knowledge of both the empirical data and collaboration theory was deliberate, ensuring that the resulting Meta-Patterns were not only reflective of the practitioner insights but also robust enough to serve as the foundation for the conceptual design work that followed. The overall process and the resulting structure enabled us to identify key areas of human-drone coordination and served as the foundation for the conceptual abstraction introduced later in the paper.

As the next step, we analyzed the identified interaction patterns in greater detail and developed them into Drone Collaboration Patterns that reflect operational use cases for human-drone interaction in emergency services. Building on the logic of AI-ThinkLets, we formulated comprehensive collaboration structures that specify all necessary components for repeatable deployment, including instructions, setup, environmental requirements, drone capabilities, and scripts for both human actors and drones. We focused on core operational use cases, such as situational monitoring, reconnaissance, communication with bystanders, logistical support, and tactical negotiation.

### *Conceptual Grounding: Toward Design Abstractions*

Following the principles of Design Science Research (Hevner et al., 2004; J. S. Lee et al., 2011), our study aims to generate both instance-level knowledge and an abstract design contribution. DSR emphasizes the dual objectives of addressing real-world problems through concrete solutions while simultaneously contributing to generalizable design knowledge that advances the research domain. In line with the framework proposed by Lee et al. (2011), our work differentiates between instance solutions, context-specific artifacts developed to solve particular problems, and abstract solutions, generalized constructs derived from instances that can inform broader classes of problems.

*Instance Solution*: The Interaction Patterns and Meta-Patterns developed in this study serve as empirically grounded, field-tested descriptions of human-drone collaboration in emergency services. These patterns synthesize practical insights gained from real-world deployments, including four field trials and 95 interviews. They represent solutions to coordination problems faced in integrating autonomous drones into ES workflows and offer a concrete foundation for understanding recurrent challenges and effective practices.

*Abstract Solution*: Building upon these instance solutions, we suggest the concept of *DroneLet* as a design abstraction. The *DroneLet* formalizes the structure, roles, and coordination mechanisms necessary for embedding autonomous drones into collaborative workflows. It extends the logic of ThinkLets and AI-ThinkLets by adapting it to drone agents' embodied, spatially situated nature.

This layered methodology ensures that our theoretical contributions are rigorously grounded in empirical practice while advancing the conceptual toolkit of CE to accommodate embodied digital agents. It bridges the problem space of human-drone coordination challenges with a solution space structured by design principles, reflecting the iterative, dual-focused nature of robust DSR.

## Results: Instance solution

Our empirical research confirms that coordination issues are central to the effective integration of drones in ES workflows. While participants consistently recognized the benefits of drone use, they also reported confusion over drone behavior, uncertainty about roles and responsibilities, and breakdowns in communication between the drone operator and the rest of the team. Even in controlled test environments, collaboration with drones often depended on improvisation, with few structured protocols to guide interaction. These insights highlight a critical gap: while the technological capabilities of drones continue to evolve rapidly, the collaborative structures required to integrate them smoothly into ES operations have not kept pace. Our paper attempts to address it by (a) providing an overview of potential scenarios in which drones can be used that demand high human-drone coordination and collaboration and (b) providing DroneLets, i.e., a concept that structure and systematize collaboration with drones while illustrating it below.

### *Drone Collaboration Pattern Catalogue*

To gain a deeper understanding of how autonomous drones can be effectively integrated into ES operations, we identified a diverse set of collaboration scenarios in which autonomous drones could play an active and meaningful role. Each collaboration scenario describes a specific instance of human-drone collaboration, typically involving a defined task, a set of actors, and a temporal or spatial dynamic that requires coordination. For example, in one scenario, a drone autonomously scans an area or a building to detect smoke or indicators of combustion and transmits this information to human responders on the ground. In a different scenario, a drone communicates with the human actors on the ground, providing guidance or it supports the operation with illumination. We derived a structured catalogue that provides a comprehensive overview of how autonomous drones can augment human emergency response activities. Table 1 summarizes the details of the Meta-Patterns and their associated Interaction Patterns.

| **Meta-Pattern** | **Description** | **Interaction Patterns** |
|---|---|---|
| Situational Monitoring | Drones observe the environment, provide live visuals, scan areas, or collect external information from surroundings or people. | - Performing quick preliminary reconnaissance before the arrival of the team<br>- Streaming real-time video footage to first responders<br>- Detecting smoke or indicators of combustion<br>- Detecting the source or location of a fire<br>- Conducting aerial surveys of the environment<br>- Conducting visual inspections through windows<br>- Entering interior spaces for close-range reconnaissance<br>- Searching an area for specified targets<br>- Collecting information from bystanders or victims<br>- Monitoring the fire site after the fire is extinguished |
| Environmental Assessment | Drones perform evaluative tasks such as analyzing fire behavior, identifying hazardous materials, checking structural safety, or locating hot spots. | - Assessing the severity and spread of a fire<br>- Analyzing smoke to determine fire characteristics<br>- Analyzing hazardous materials container markings or placement<br>- Detecting the presence of hazardous materials<br>- Evaluating air safety by entering recently affected structures/areas<br>- Locating hot spots in the affected area<br>- Searching for hydrants or other critical infrastructure<br>- Identifying the presence and location of victims or bystanders<br>- Evaluating victim health status through sensor analysis |
| Information Exchange and Guidance | Drones guide or inform human actors, broadcast messages, or relay communication between bystanders and first responders. | - Providing verbal or visual instructions to first responders<br>- Broadcasting information to nearby bystanders or victims<br>- Responding to human attempts to initiate communication with the drone<br>- Relaying communication from bystanders or victims to response teams<br>- Providing directional guidance to first responders<br>- Guiding bystanders toward first responders |
| Aerial Delivery Support | Drones deliver physical objects (rescue kits or fire-extinguishing capsules). | - Delivering general-purpose supplies or tools<br>- Transporting rescue kits to injured or isolated individuals<br>- Deploying fire-extinguishing payloads to targeted areas |
| Suspect Engagement and Tracking | Drones assist in locating, pursuing, negotiating with, or monitoring suspects in police operations. | - Conducting remote negotiation with suspects<br>- Persuading suspects to comply or surrender<br>- Searching for suspects within the operational area<br>- Pursuing suspects across terrain or obstacles<br>- Marking suspects with visual indicators for tracking<br>- Applying pressure through persistent observation or presence<br>- Gathering situational data on suspects' actions or positions<br>- Deterring suspects through visible or audible presence |
| Drone-to-Drone Coordination | Drones collaborate with other drones to support joint missions. | - Issuing commands or task assignments to other drones |

| Geospatial Intelligence Support | Drones enhance situational awareness by marking mission-relevant features and verifying address locations. | - Annotating environmental features or findings on a digital map<br>- Verifying addresses or locations relevant to the operation |
|---|---|---|
| Personnel Transport | Drones transport first responders or personnel to or from operational areas. | - Transporting first responders to or from operational areas |
| Operational Infrastructure Support | Drones improve operational environments by offering lighting or communication access points. | - Acting as a mobile signal or communication relay point<br>- Providing illumination in low-visibility or nighttime conditions<br>- Offering short-range digital communication (e.g., Bluetooth) to people on site |
| Operational Status Reporting | Drones communicate their own status, location, or progress to the human team during operations. | - Communicating drone status and operational updates to team members |

**Table 1. Human-drone collaboration patterns in emergency operations**

The largest groups – Situational Monitoring and Environmental Assessment – reflect the central role of drones in enhancing situational awareness and evaluating environmental conditions. Drones observe dynamic scenes, gather critical information, and perform analyses such as detecting hazards, identifying structural risks, or assessing victim conditions. Other important categories include Information Exchange and Guidance, where drones actively interact with human actors by guiding, relaying communication, or broadcasting information, and Aerial Delivery Support, where drones take on logistical tasks by transporting essential supplies or firefighting equipment.

In more tactical and law enforcement contexts, drones contribute to Suspect Engagement and Tracking, supporting operations through surveillance, negotiation, or active pursuit. Additional meta-patterns, such as Drone-to-Drone Coordination and Geospatial Intelligence Support, highlight how autonomous systems can collaborate with each other or contribute to mission-critical mapping activities.

Specialized roles such as Personnel Transport, Operational Infrastructure Support, and Operational Status Reporting further illustrate the breadth of potential drone involvement, ranging from physical transportation of responders to providing illumination, communication access points, or real-time status updates.

## *"SkyMessenger" – Enhancing Awareness Among Bystanders and Victims*

Collaboration within the identified interaction and meta-patterns is rarely straightforward. These patterns often involve multiple human and non-human actors, such as firefighters, drone officers, autonomous drones, and digital systems, who coordinate their actions to achieve a shared goal. Successful execution depends not only on shared understanding and timing, but also on a well-prepared physical and digital environment, reliable communication infrastructure, and clearly defined procedures. In practice, however, these conditions are often difficult to establish and maintain in high-stakes, dynamic emergency settings.

To systematically support these collaborations, we designed a set of Drone Collaboration Patterns, including examples such as *SkyMessenger* (broadcasting information to victims), *DeepSeeker* (detailed site reconnaissance with specialized sensors), *WindowWatcher* (visual inspection through windows), *AreaSearcher* (systematic area scanning), and *RescueCourier* (delivery of emergency supplies), etc. These patterns provide structured and modular building blocks for engineering effective human-drone collaboration workflows across various emergency service scenarios.

| With *SkyMessenger*, the drone supports the emergency service team by transmitting the message to the persons in the defined area. | |
|---|---|
| **Choose this pattern…** | … when you need to transmit a message to victims or bystanders<br>… when victims or bystanders have been located and identified<br>… when victims or bystanders cannot yet be reached physically by responders<br>… when direct communication is delayed or impossible<br>… in chaotic or dangerous environments |
| **Do not choose this pattern…** | … if the situation requires sensitive, individualized communication (e.g., delivering bad news)<br>… if environmental noise or risks would render aerial communication ineffective |
| **Input** | - Message to be transmitted<br>- Definition of the recipient (directly or through spatial specification) |
| **Output** | - Broadcast confirmation |
| **Environment: Physical** | - A drone port with multiple drones and charging infrastructure that allows drones to take turns during each other's charging cycles is needed.<br>- Victims or bystanders must be located within a defined operational area.<br>- Open communication zones should be prioritized where visibility and audibility are sufficient for effective broadcasting<br>- Tablet with the installed Drone Operation System and fully functional microphone |
| **Environment: Digital** | - The drone must have the Drone Operation System installed, enabling responders to define or update broadcast messages dynamically.<br>- Message templates for different scenarios (e.g., evacuation instructions, reassurance messages) should be prepared in advance<br>- Text-to-voice tool |
| **Drone Capabilities** | - The drone must be equipped with loudspeakers capable of broadcasting clear and understandable messages.<br>- Ideally, it should also feature lighting signals (e.g., flashing lights) to attract attention if necessary. |
| **Drone Appearance** | The drone should be visibly associated with emergency services, using standardized colors or insignias to reinforce legitimacy and trust among victims and bystanders. |
| **Drone Setup** | - The drone must be fully charged<br>- The drone must be connected to the Drone Operation System<br>- The drone must be loaded with predefined or customizable messages<br>- Text-to-voice tool installed |
| **Environment Setup** | Ensure that the communication broadcast area is cleared of major sound obstructions (if possible) and that emergency teams are ready to intervene when needed. |
| **Script Human Actors** | 1. Select an appropriate broadcast message based on the evolving situation<br>If necessary, type or record a new message for the target group.<br>2. Assign a broadcast area or target group.<br>3. If necessary, define the time interval for the broadcast.<br>4. Send the command to the drone.<br>5. Monitor bystander or victim reactions in the Drone Operation System and decide when manual intervention or message updates are necessary. If necessary, go back to Step 1.<br>6. Coordinate positioning based on crowd movement and updated information needs. If necessary, go back to Step 1. |
| **Script Drone** | 1. Receive a broadcast command.<br>2. Launch and navigate to the respective area or location. In case of multiple recipient locations, calculate an optimal route.<br>3. In case of a text message transform it to a voice message.<br>4. Arrive on spot.<br>5. Find the best altitude for the broadcast.<br>6. Maintain position or adjust based on crowd movement patterns.<br>7. If necessary, use lighting signals to enhance visibility.<br>8. Broadcast safety messages, instructions, or reassurances at defined intervals at the designated location<br>9. Send Broadcast conformation and the footage of the bystanders/victims to the Drone Operation System.<br>10. If new information from the drone officer received go back to 1.<br>11. If the battery charge is low: Return to the drone port for recharging, order another drone to take over.<br>12. Once ordered to, return to the drone port. |
| **Error Handling** | 1. Assess and log the anomaly, exception, or error.<br>2. Attempt automated recovery for known errors (retry the transmission, recalculate the path, switch communication channels).<br>3. Notify the drone officer and escalate, stating the error type and informing that going into the state of contingency.<br>4. Execute failsafe maneuver: stop the action, ascend to a pre-set altitude, hover.<br>5. If human command arrived: execute it.<br>6. If no human command arrived and enough power: return to drone port.<br>7. If no human command arrived and not enough power for return: generate loud, unpleasant sound and attempt gradual descent on the current spot. |
| **You are done if …** | … Broadcast confirmation received<br>… No messages to broadcast |

**Table 2. The Drone Collaboration Pattern "*SkyMessenger*" – Enhancing Awareness Among Bystanders and Victims**

While the meta-patterns and interaction patterns offer a useful taxonomy of potential drone applications, they operate at a high level of abstraction. To show how these concepts translate into concrete, field-ready procedures, we present a detailed example. The Drone Collaboration Pattern called SkyMessenger demonstrates this translation. It bridges theory and practice by showing how the abstract logic becomes a specific, repeatable, and structured workflow. This example represents a typical case that shows practitioners how a DroneLet converts the complex coordination between human actors, the drone, and the operational environment into an actionable script. This pattern, situated within the meta-pattern of Information Exchange and Guidance, focuses on the use of autonomous drones to broadcast critical information to bystanders and victims during emergency situations. It exemplifies how collaboration extends beyond the immediate operational team to include affected civilians, requiring careful coordination between human actors and autonomous systems. The collaboration activities, setup requirements, and procedural steps involved in this pattern are detailed in Table 2.

# Discussion

Our study addressed two research questions. We answered RQ1 by systematically analyzing field trial interviews, resulting in a catalogue of 44 Interaction Patterns grouped into 10 Meta-Patterns. These patterns reflect how practitioners envision the integration of drones into operational workflows. The results demonstrate that collaboration with drones in ES is both promising and complex. The identified interaction patterns reveal that such collaboration frequently involves multiple actors and real-time coordination across physical and digital environments. While these concrete patterns offer valuable insight into how drones can be embedded in operational workflows, they also highlight the need for generalized, reusable design elements that can structure and support this form of collaboration across different contexts and scenarios.

Current approaches to deploying drones often rely on ad hoc setups, trial-and-error practices, or implicit knowledge held by experienced operators. In many emergency service organizations, drone use is either outsourced to specialized external units that are called in when needed or handled internally by dedicated drone pilots who operate the systems based on their own expertise. While both models offer flexibility, they also introduce limitations: they are highly dependent on individual experience, lack standardized procedures, and may be difficult to scale or coordinate across organizational boundaries. In high-stakes environments, such reliance on informal practices can lead to coordination breakdowns, delays, and missed opportunities for integrating drones more effectively into the broader workflow.

A thoughtful and planned approach for interaction with drones aligns with ES dependency on scripts and processes expressed, for instance, as SOPs (Butler et al., 2023; Duncan et al., 2014; Weinschenk et al., 2008). Even though some situations require decision discretion and improvisation, those remain exceptions and take standardized procedures as a departure point (Carvalho et al., 2018; Jahn, 2016; Toups Dugas & Kerne, 2007). Currently, the use of drones in firefighting and policing relies on improvisation or too abstract guidance detached from specific application scenarios, which invariably transforms otherwise standard situations with adequate existing procedures into extraordinary ones. This increases safety risks and demands on all participants. Therefore, to implement agentic systems like autonomous drones at scale, existing SOPs will require a significant update to accommodate this new type of entity. The catalogue presented in Table 1 offers an empirically supported and structured list of potential extensions to existing SOPs. We speculate that the presented Meta-Patterns might be instantiated as stand-alone procedures for specific situations or integrated into existing scripts where appropriate. We invite chiefs of fire and police departments to assess the relevance of the identified patterns for their specific contexts and investigate how they can be used to extend their procedures and protocols. Considering DroneLets as a standard for specifying scripts might be effective in ensuring that important decisions and behaviors are codified.

Collaboration with drones is inherently complex. It requires coordination across physical, digital, and organizational layers. Achieving effecting collaboration is even more complicated due to the nature of ES. Autonomy may offer a promising solution to these challenges by enabling drones to perform certain tasks independently. To systematically address these challenges and address RQ2, we propose an abstract solution that formalizes collaboration design for autonomous drone use: *DroneLets*. Building on the lineage of ThinkLets and AI-ThinkLets (Schwabe et al., 2025), *DroneLets* offer a structured and repeatable framework for engineering human-drone interaction, tailored to the needs of ES. *DroneLets* should also minimize cognitive load by limiting human decision points (e.g., automated message selection) and include training to build trust in drone autonomy. In the following, we outline the core components of the *DroneLet* concept and illustrate how it captures the necessary setup, capabilities, and coordination logic to support embodied, autonomous agents in collaborative tasks.

## *Abstract solution: DroneLets*

The concept of *DroneLets* builds on a rich tradition of structuring collaboration through repeatable design elements. Originally, ThinkLets were introduced as compact, reusable collaboration techniques that support practitioners in achieving specific group goals (Briggs et al., 2003). With the rise of AI-based collaboration tools, the AI-ThinkLet extended this logic to hybrid teams that include digital agents, LLMs, capable of participating in structured collaboration (Schwabe et al., 2025). AI-ThinkLets assume that all collaboration occurs in a digitally bounded space and thus requires an expansion of the original framework to include not only human activities but also the functionality, configuration, and integration of the digital agent itself. This included new dimensions for setup (e.g., agent context) and script (e.g., agent behavior in dialogue), as well as considerations for language style, contribution frequency, and tone adaptation. However, drones are embodied agents that operate in dynamic, real-world physical environments. Effective collaboration with drones requires not only coordination of information exchange but also careful consideration of hardware characteristics, physical positioning, environmental constraints, and safety-critical setups.

The DroneLet construct is admittedly more complex than its predecessors. This increased complexity, however, stems from the real challenges of moving from purely digital collaboration to working with embodied agents in dynamic, safety-critical physical environments. Each component, from 'Drone Appearance' to the physical 'Environment' setup, captures the necessary details to ensure collaboration processes are repeatable, safe, reliable, and effective in real-world applications.

As autonomous and embodied agents, drones become increasingly integrated into operational workflows, making further conceptual development necessary. The *DroneLet* represents the next evolution in this lineage, expanding the ThinkLet family to accommodate the complexity of human-drone collaboration. The real-world elements, such as drone capabilities (e.g., sensors, payloads), environmental preparations (e.g., drone ports, flying zones), and operational readiness (e.g., battery management), are absent from AI-ThinkLets. To bridge this gap, we extended the AI-ThinkLet concept and developed *DroneLets*, which systematically capture both the digital coordination and the physical embodiment aspects necessary for structuring collaboration with autonomous drones.

Figure 1 illustrates this progression – from ThinkLets to AI-ThinkLets to DroneLets – by visualizing the progressive expansion of the ThinkLet framework to accommodate digital agents and embodied agents such as drones (DroneLets). Each stage introduces additional components to address the capabilities, setup, and coordination requirements of increasingly complex collaboration settings.

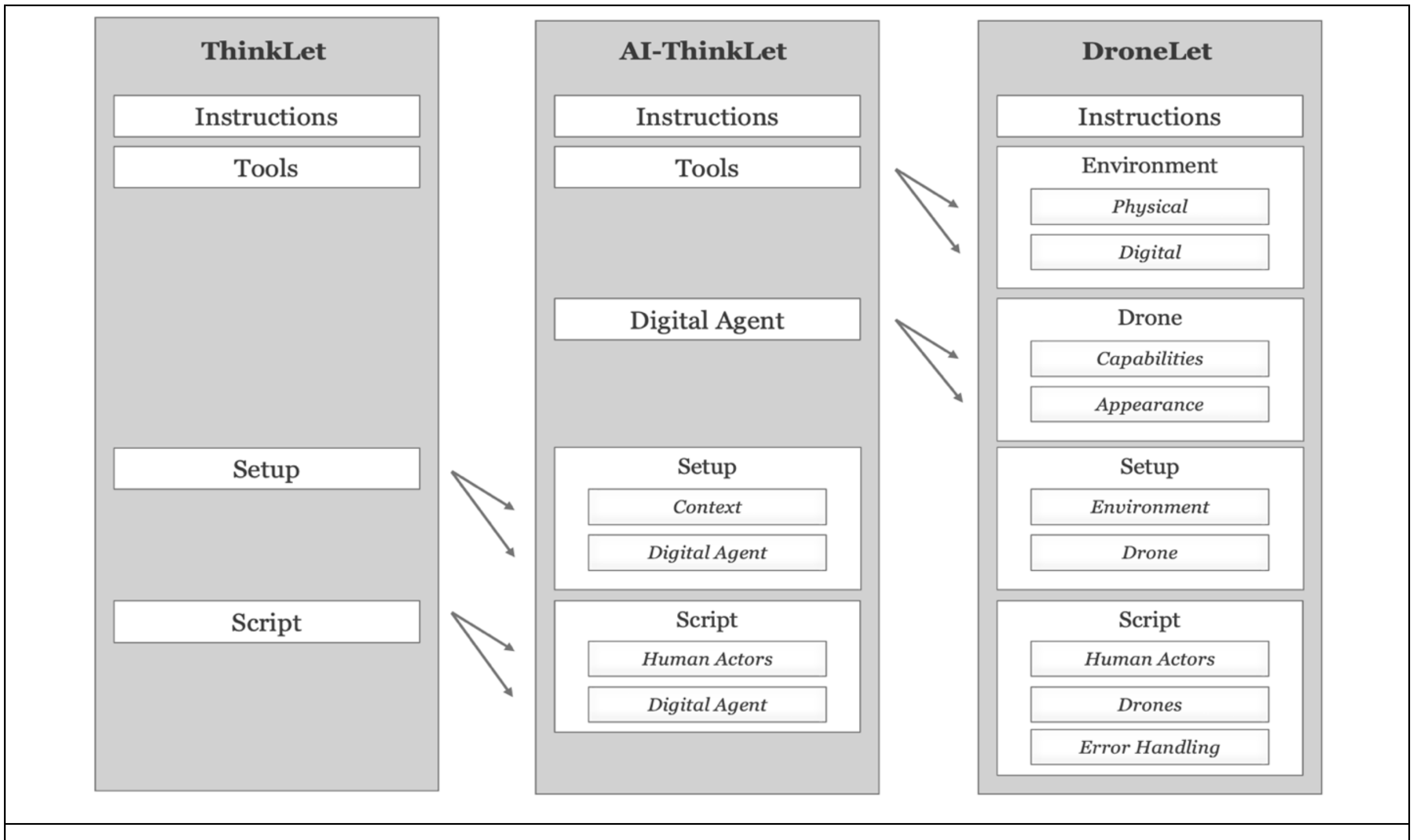

**Figure 1. Evolution of Collaboration Design Constructs: From ThinkLets to AI-ThinkLets to DroneLets (based on Schwabe et al. (2025))**

The *DroneLet* expands the traditional ThinkLet structure by eight new components, each reflecting a critical aspect of collaboration with autonomous, embodied agents:

*Instructions,* as in ThinkLet*,* define when the *DroneLet* should (or should not) be applied. It provides guidance on suitable use cases, operational constraints, and environmental conditions under which the collaboration pattern is valid.

The *Environment* component defines the conditions and resources in the physical and digital surroundings that are necessary for enabling effective human-drone collaboration. It includes both infrastructure and software systems that support communication, monitoring, and operational readiness.

- *Physical*: Specifies the physical infrastructure or artifacts needed, such as a drone port, charging equipment, communication gear, or tablets or other portable devices.
- *Digital*: Covers digital tools and systems, including drone control software, real-time alert platforms, communication interfaces, and map or sensor data integrations.

The *Drone* components specify the technical and perceptual properties of the drone that are required for the pattern. It covers both the drone's functional capabilities and its appearance, which may influence user trust, recognizability, or social acceptance.

- *Capabilities*: Describes what the drone must be able to do, including sensory equipment (e.g., thermal cameras, gas detectors) and mobility features.
- *Appearance*: Addresses visual or auditory design elements (e.g., color, shape, lights, logos) that influence human trust, recognizability, or compliance in public or chaotic environments.

The *Setup components* outline the preparatory actions required to enable collaboration. It includes both configuring the environment and preparing the drone, ensuring all elements are in place before the operational phase begins.

- *Environment*: Outlines preparatory steps to make the physical and digital environment ready for collaboration (e.g., clearing space, activating systems).
- *Drone*: Involves preparing the drone itself – charging, configuring the mission area, updating firmware, and loading digital tools or communication scripts.

The *Scripts* define the collaborative sequence of actions for both human actors and the drone. It ensures role clarity, timing, and coordination throughout the operation, supporting predictable and repeatable collaboration dynamics.

- *Human Actors*: Specifies what human participants (e.g., drone officer, incident commander) must do, including initiating commands, monitoring status, or reacting to alerts.
- *Drones*: Details the drone's behavior during the collaboration, such as patrol paths, triggering logic, communication routines, or handover conditions.
- *Error Handling*: Details the drone's behavior in case of anomalies, exceptions, and errors offering a backdrop procedure which is highly relevant due to intermediate physical danger that a randomly acting drone can cause; this is particularly relevant in emergency services where safety and feeling of safety are of high importance.

This component model allows *DroneLets* to function as modular collaboration designs, each encoding a reusable solution to a specific type of human-drone coordination problem. Just as ThinkLets facilitated structured human interaction and AI-ThinkLets enabled human-AI dialogue, *DroneLets* now offer the tools to engineer predictable and effective collaboration between humans and autonomous drones in real-world operational environments.

This work contributes to the field of Collaboration Engineering by introducing a novel artifact type. While previous research has explored AI-ThinkLets for virtual agent integration, *DroneLets* advance the framework by addressing the spatial, sensorimotor, and coordination demands of drones operating in dynamic physical environments. *DroneLets* offer a design toolkit for practitioners seeking to integrate autonomous drones into ES workflows in a repeatable, safe, and structured manner. By formalizing collaboration patterns as reusable design elements, they support incident planners, drone officers, and system and hardware designers in anticipating and engineering effective human-drone interactions. This has direct relevance for improving operational reliability, role clarity, and coordination efficiency in high-stakes environments.

The modular design of *DroneLets* not only enables structured human-drone collaboration in specific operational scenarios but also offers flexibility for future adaptations as technology evolves. One potential development concerns the merging of currently distinct *DroneLets*. In the current framework, some *DroneLets* are separated due to technological constraints, such as differences in payload requirements, sensor configurations, or specialized equipment. Keeping these functions separate helps optimize drone performance by minimizing weight, reducing operational complexity, and ensuring rapid deployment. However, as technological advances lead to lighter, more integrated equipment, it may become feasible to merge previously distinct *DroneLets* into multifunctional patterns without compromising efficiency or speed. The *DroneLet* framework is therefore intentionally designed to be adaptable to ongoing technological progress.

Beyond merging, sequential execution of multiple *DroneLets* with the same set of drones represents another promising direction. While the current *DroneLets* are atomic and independent, emergency workflows often involve phases where tasks naturally follow one another. It would be advantageous for drones to transition seamlessly from one collaboration pattern to the next – such as moving from an initial reconnaissance task to a communication or monitoring role – without requiring full redeployment or reconfiguration.

However, sequential execution increases the demands on drone capabilities, environment setup, and coordination processes. It requires drones to be equipped with the combined functionalities of multiple collaboration patterns, and the environment must support these extended operational phases. Given current technological limitations – particularly in payload, endurance, and autonomous adaptability – such sequential operations must be planned in advance to ensure feasibility and effectiveness.

The development of *DroneLets* is grounded in a progression from real-world coordination challenges observed in drone deployment to the formulation of structured design solutions. Starting from field trials and interviews, we identified recurring patterns of collaboration, distilled them into generalized structures, and abstracted them into a formal design concept. In this way, *DroneLets* not only address the complexity of present-day collaboration needs but also provide a design framework for structuring human-drone interaction as technologies and operational demands continue to evolve.

Beyond contributing to the CE discourse, DroneLets directly assist practitioners, such as chief officers in firefighting and police departments. When revising procedures and scripts for their respective organizations, they can use DroneLets as guidance to establish specific, actionable procedures for collaboration with drones. The structure imposed by DroneLets acknowledge the unique autonomy, capabilities, and appearance of drones, as well as the hybrid physical and digital nature of interacting with them. The DroneLet example in Table 2 is exemplary for a drone application that we anticipate will become one of the most common. Practitioners can directly integrate this DroneLet into their SOPs. Given the significant differences in SOPs between countries and departments, using DroneLets could be the first step in harmonizing standards, at least by implementing a shared structure that supports the transferability of scripts across units. With this in mind, we acknowledge that establishing standard procedures for drone collaboration is a long-term endeavor, and this manuscript represents only a first step in that direction. The proposed approach can help transform drone use from an improvised, specialist task into a standardized, repeatable capability, ensuring that drone operations are a seamless part of ES workflows.

While the DroneLet framework offers a solid foundation for human-drone collaboration, full-scale implementation encounters substantial real-world constraints. From a technological standpoint, complex DroneLets need major advances in several areas: extending battery life for longer missions, improving sensor reliability for accurate environmental assessment, and developing autonomous navigation algorithms for cluttered or GPS-denied environments. Regulatory challenges prove equally demanding. Current aviation regulations in many regions, including Switzerland, strictly limit autonomous beyond-visual-line-of-sight operations, particularly in populated areas. These rules typically require human pilot intervention capability, which directly constrains deployable autonomy levels.

## Conclusion and Future Work

This study extends the principles of Collaboration Engineering to the emerging challenge of structuring collaboration between humans and autonomous, embodied systems – specifically drones operating in high-stakes ES environments. Building on the evolution from ThinkLets to AI-ThinkLets, we introduced DroneLets as a novel artifact type that formalizes repeatable collaboration patterns between human teams and autonomous drones.

Grounded in four field trials and 95 participants interviews with first responders, police officers, and bystanders, we developed a catalogue of 44 interaction patterns, grouped into 10 meta-patterns, from which generalized design principles were abstracted. By detailing examples such as *SkyMessenger*, we demonstrated how *DroneLets* can structure human-drone workflows beyond ad hoc practices. These patterns illustrate how drones can be integrated not merely as tools, but as collaborative agents within structured workflows – requiring appropriate setup, role clarity, and coordination logic.

While this research offers a significant conceptual contribution, several limitations should be acknowledged. First, this research is situated within the Swiss organizational and regulatory context. Future research is needed to validate and adapt these findings in diverse global settings. Whereas many European countries like Germany or Poland implement a similar model of fire prevention highly depending on local, yet well-trained volunteering brigades, countries from other parts of the world might need to adapt the observations of the current study to their local circumstances. Similarly, in Switzerland, like in most European countries, the legal framework surrounding the operational use of drones by emergency services is still evolving. Although we drew on the current Swiss regulatory landscape, changes to the legal environment are likely, and future research may expand this perspective to a broader international context.

Second, while the interaction patterns and DroneLets were grounded in rich empirical data, they were not directly validated with police officers, firefighters, or field practitioners during the design process. The selection, refinement, and abstraction of collaboration patterns were based on interview analysis rather than practitioner co-design. Future work should involve direct engagement with first responders, ensuring that *DroneLets* align closely with practical operational needs and domain expertise.

Third, the *DroneLets* proposed in this study were not implemented or evaluated in operational trials. Given the current technological limitations of drone autonomy, many *DroneLets* anticipate capabilities that are still under development. Once autonomous drones reach higher levels of maturity, future field trials will be necessary to validate, refine, and evolve the proposed collaboration structures.

Future research should focus on broader validation of the *DroneLet* concept across additional domains such as disaster response, civil protection, or industrial inspection. This includes not only expanding the repertoire of *DroneLets*, but also testing their generalizability and adaptability in different organizational and technological contexts. This might involve more extended exchange with experts and practitioners.

Furthermore, to accelerate the adoption and practical application of this research, future work should focus on developing open-source DroneLet *toolkits and templates* to help scale the solution proposed here. These resources would provide emergency service organizations with a practical starting point for adapting and implementing these patterns. A toolkit could include customizable script templates, checklists for the 'Setup' and 'Environment' components, and guidance for aligning DroneLets with local regulations and existing SOPs. By lowering the barrier to entry, such resources would empower practitioners to move beyond theoretical models and begin engineering their own context-specific, repeatable human-drone collaboration processes, thereby fostering a community of practice around this emerging field.

Finally, as technology advances, the concept of *DroneLets* could be extended even further. Future research may explore the creation of RobotLets – collaboration structures for a wider variety of embodied autonomous agents, such as service robots, autonomous vehicles, or rescue robots. This would broaden the applicability of Collaboration Engineering principles to an even larger class of hybrid human-machine teams, preparing CE for the challenges of an increasingly autonomous and embodied digital future.

## Acknowledgment

This research was a part of the project titled 'Automated Decision-Making and the Foundations of Political Authority,' financially supported by the Swiss National Science Foundation (project number 208013). We would like to convey our sincere appreciation to all project members for their constructive feedback and active participation throughout the development and evaluation phases of the drone system. Special acknowledgment is extended to the participants in the field trials for generously sharing their insights and feedback. Our gratitude goes to the Altdorf and Silenen fire stations and the dedicated firefighters involved in the field trials for their steadfast support and collaboration during the exercises. We would also like to thank the University of Zurich members – Andreas Bucher, Sven Eckhardt, Yiwei Wu, Beat Furrer, Simon Giesch, Yiming Hu, Yinglun Liu, Moyi Li and Pascal Marty – for their indispensable assistance in data collection and analysis. Furthermore, we express our thanks to the anonymous participants in our study and the review team for their invaluable insights and guidance in refining this manuscript.